\documentclass{article}

\usepackage{PRIMEarxiv}

\usepackage[utf8]{inputenc} % allow utf-8 input
\usepackage[T1]{fontenc}    % use 8-bit T1 fonts
\usepackage[breaklinks=true,bookmarks=false]{hyperref}       % hyperlinks
\usepackage{url}            % simple URL typesetting
\usepackage{booktabs}       % professional-quality tables
\usepackage{amsfonts}       % blackboard math symbols
\usepackage{nicefrac}       % compact symbols for 1/2, etc.
\usepackage{microtype}      % microtypography
\usepackage{lipsum}
\usepackage{fancyhdr}       % header
\usepackage{graphicx}       % graphics
\title{Prevention is better than cure: a~case study of the abnormalities detection in~the~chest}

\author{Weronika Hryniewska \thanks{These authors contributed equally: Weronika Hryniewska, Piotr Czarnecki, Jakub Wiśniewski \newline Work on this paper was funded by the IDUB against COVID-19 initiative at~the~Warsaw University of~Technology.} $ ^1$\\
{\tt\small w.hryniewska@mini.pw.edu.pl}
\And
Piotr Czarnecki $^*{}^1$\\
{\tt\small 	piotr.czarnecki3.dokt@pw.edu.pl}
\And
Jakub Wiśniewski $^*{}^1$\\
{\tt\small 	jakub.wisniewski10.stud@pw.edu.pl}
\AND
Przemysław Bombiński $^2$\\
{\tt\small 	przemyslaw.bombinski@uckwum.pl}
\AND
Przemysław Biecek $^1$\\
{\tt\small 	przemyslaw.biecek@pw.edu.pl} \\ 
[2ex]
\normalsize
    $^1$Faculty of Mathematics and Information Science, Warsaw University of Technology\\%
\normalsize
    $^2$Department of Pediatric Radiology, Medical University of Warsaw\\%   
}

\begin{document}
\maketitle

\begin{abstract}
    Prevention is better than cure. This old truth applies not only to the~prevention of diseases but also to the~prevention of issues with AI models used in medicine. The source of malfunctioning of predictive models often lies not in the~training process but reaches the~data acquisition phase or design of the~experiment phase. 
    
    In this paper, we analyze in detail a~single use case - a~Kaggle competition related to the~detection of abnormalities in X-ray lung images.  We demonstrate how a~series of simple tests for data imbalance exposes faults in the~data acquisition and annotation process.  Complex models are able to learn such artifacts and it is difficult to remove this bias during or after the~training. Errors made at the~data collection stage make it difficult to validate the~model correctly.
    
    Based on this use case, we show how to monitor data and model balance (fairness) throughout the~life cycle of a~predictive model, from data acquisition to parity analysis of model scores. 
\end{abstract}

%%%%%%%%% BODY TEXT
\section{Introduction}

Radiography (X-ray) is an~imaging technique that uses a~small dose of ionizing radiation to create images of the~internal structures of a~body. Due to the~relatively low price of the~device and the~existence of portable devices, X-ray imaging is a~widely used technique. However, it is particularly difficult to assess the~severity of~the~pathology, and, thus, only experts in radiology should interpret chest images.

Recent applications of machine learning (ML) have gained popularity in the~medical domain \cite{Yu_2020_CVPR, Guo_2020_CVPR}. The performance achieved by neural networks is becoming similar to that reached by medical experts \cite{Zhou_2019_CVPR}.

Considering the~need for a~highly precise and~fast diagnosis process, on the~Kaggle platform was announced a~competition about automatically localizing and classifying thoracic abnormalities from chest radiographs \cite{Nguyen2020}. On December 30, 2020, the~database with 18,000 posterior-anterior (PA) X-ray scans in DICOM format became available on: \cite{VingroupBigDataInstitute2020}. More than 1,300 teams are participating in the~competition trying to train the~best model. The total prize money in this challenge is 50,000 dollars. The crucial value of the~dataset is in the annotations. They were created by radiologists and show the location of anomalies in chests.

Although a~great deal of work was put into creating the~database and annotating the~chest images, there were many difficulties and problems which influence training reliable neural network models. 

In this paper, we will discuss the~problems that are present in the~training set, and show possible solutions. We believe that the~criticism, together with the~proposed improvements, will contribute to building better datasets and then creating more reliable models for localization and classification tasks.

%------------------------------------------------------------------------

\section{Problems in the~training set}
The training set contains 15,000 lung images in DICOM format with annotations. Each image was annotated by three radiologists. Due to a~DICOM format, the~images have high quality and the~information about a~patient (such as age or sex) or about the~image (such as the~number of allocated bits) is included.

There are fourteen labels for lesions and one additional label for images of healthy lungs: Aortic enlargement, Atelectasis, Calcification, Cardiomegaly, Consolidation, ILD, Infiltration, Lung Opacity, Nodule/Mass, Pleural effusion, Pleural thickening, Pneumothorax, Pulmonary fibrosis, Other lesions, No finding.

The test set has 3,000 DICOM files with no annotations as the~challenge is ongoing.

%------------------------------------------------------------------------

\begin{figure}[!h]
    \centering
    \includegraphics[width=0.49\linewidth]{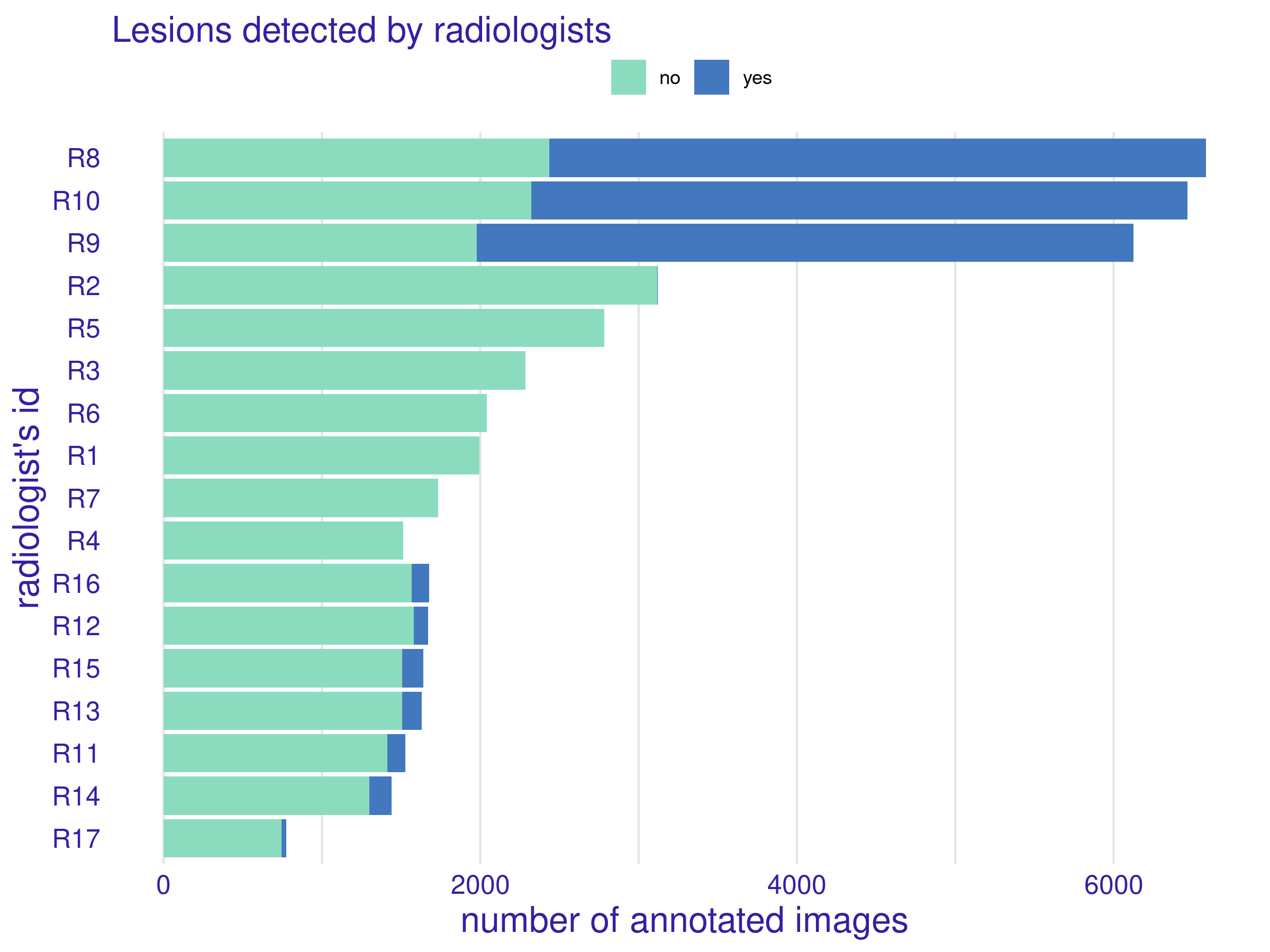}
    \includegraphics[width=0.49\linewidth]{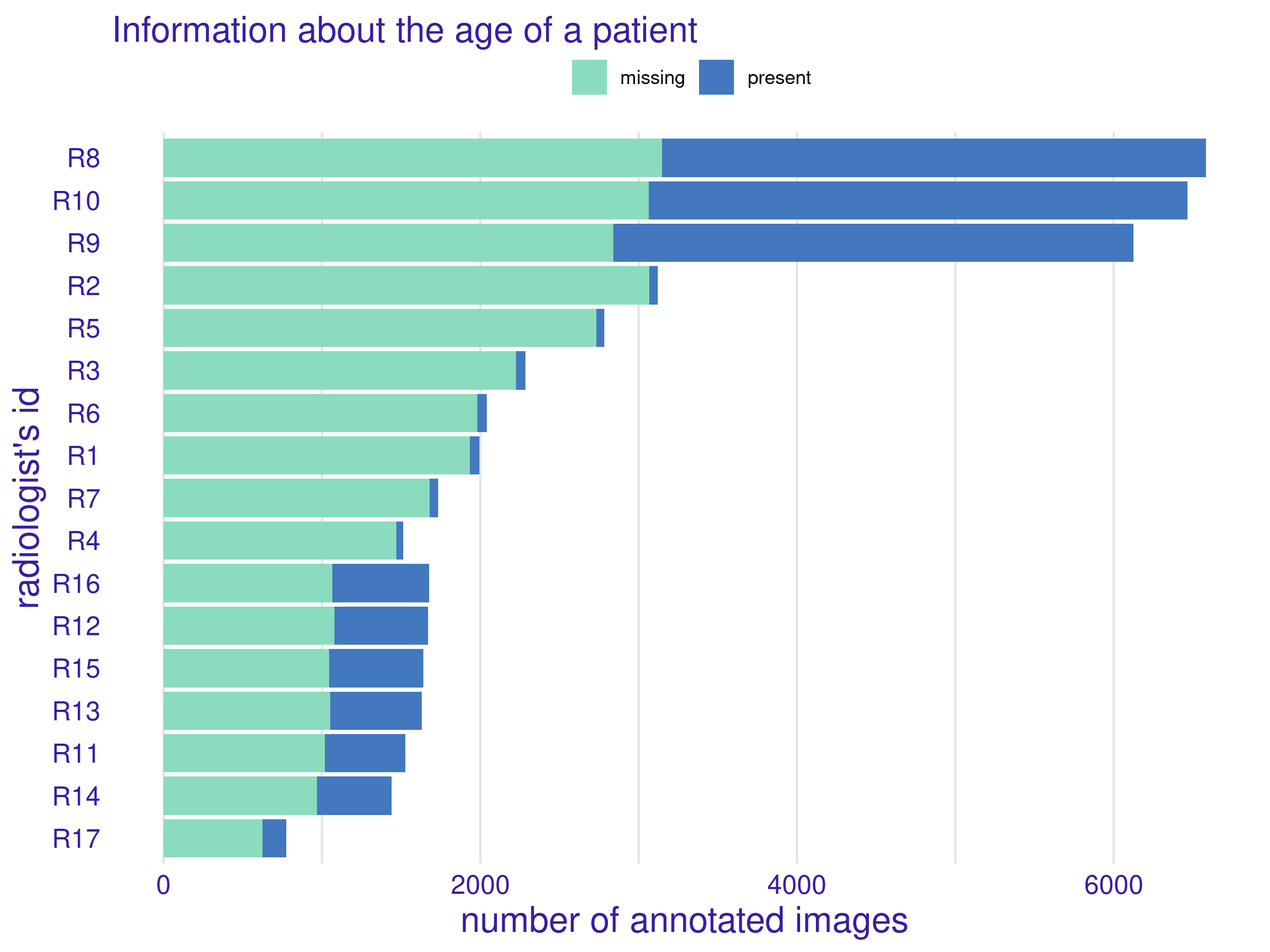} \\
    \includegraphics[width=0.49\linewidth]{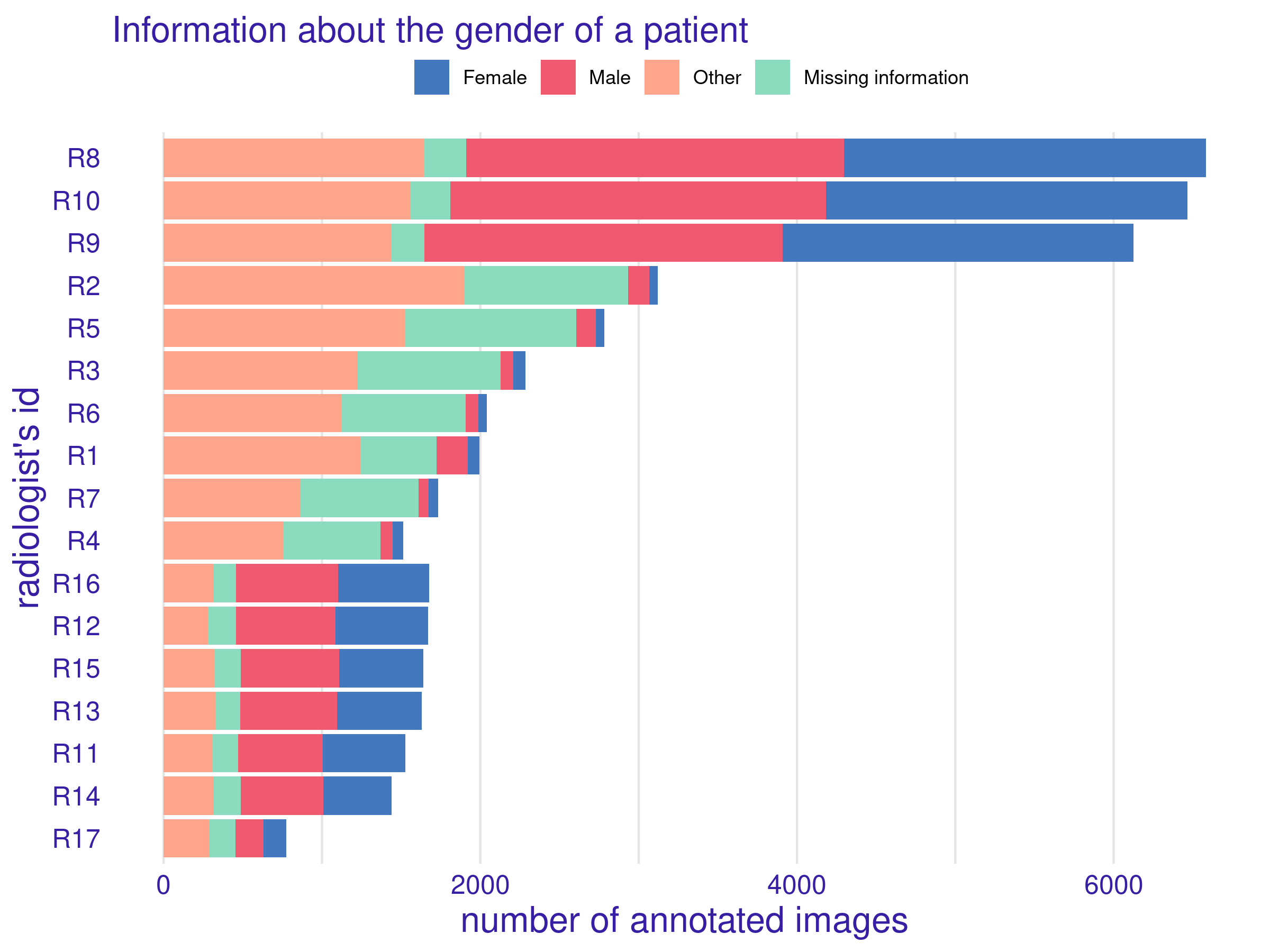}
    \caption{Different label distributions among radiologists. The top left plot shows the~number of images annotated by a~radiologist grouped by whether the~illness was found or not. The top right plot shows grouping by age and the~bottom one grouping by sex.}
    \label{fig:labels}
\end{figure}

\subsection{Consistency among radiologists}

\paragraph{\textbf{Unequal division of annotation work between radiologists}} As visible in Figure \ref{fig:labels}, the~radiologists can be divided into three groups. 

The first group, R8-R10 worked on the~same part of the~X-ray dataset and annotated most of the~images present in the~dataset, both images with and without findings. Each radiologist annotated more than 6,000 images. Those three radiologists annotated 95\% of all of the~detected findings in this dataset.

The next group R1-R7 did not detect almost any lesion (R2 found 3, the~rest none). In addition, in this group, information about age is missing in the~vast majority of cases. The information about gender is either set to "other" or missing.

The last group, R11-R17. Each radiologist annotated less than 2,000 images with a~high fraction of 'no findings' images. However, in most cases information about gender is present.

We suggest that radiologists ought to be assigned randomly to the~images. Annotations that already exist should not be shared between radiologists.

\paragraph{\textbf{Not clear annotation rules}} 
Comparing the~class labels given by different radiologists for a~particular image, the~consistency is remarkably low. In Table \ref{tab:consistency}, in group R8-R10 (radiologists that annotated 95\% of all findings), radiologists agreed with both colleagues on all classes only in 46\% of images.

\begin{table}[!h]

\centering

\caption{Radiologists annotations consistency.}
\begin{tabular}{|p{7.2cm}||p{1.0cm}|p{1.2cm}|p{1.3cm}|}
\hline
Radiologists & R1-R7 & R8-R10 & R11-R17 \\
\hline\hline
Agreed with at least one colleague on all classes & 100\% & 69\% & 96\% \\
\hline
Agreed with both colleagues on all classes & 100\%  & 46\% & 94\% \\
\hline
\end{tabular}
\label{tab:consistency}
\end{table}

It is worth mentioning that all annotators are radiologists with at least 8 years of experience. For this reason, we assume the~low consistency could be caused by not clear annotation instructions in general. Due to the~fact that one of us has a~specialization in radiology, we found an~explanation.
Some anomalies may be typical for specific age - like aortic enlargement. However, some radiologists mark it as an~anomaly, while others ignore it as an~acceptable finding for a~patient of this age.

For ML problems, in the~training set, anomalies should be marked consequently, no matter if it is a~typical anomaly for patient age. For this reason, it seems crucial to control the~annotation process and clarify annotation rules. This could be a~role of an~expert radiologist.

\paragraph{\textbf{Different label for the~same pathology}} Another effect of unclear annotation rules is the significantly overlapping definitions of anomalies. A~class ILD and Pulmonary fibrosis strongly overlap, similarly to Consolidation and Infiltration. The most vivid example is a~"lung opacity", which covers six other classes! It will cause obvious inconsistency, if not clearly stated at the~beginning of the~annotation process.

\begin{figure}[!h]
    \centering
      \includegraphics[height=5cm]{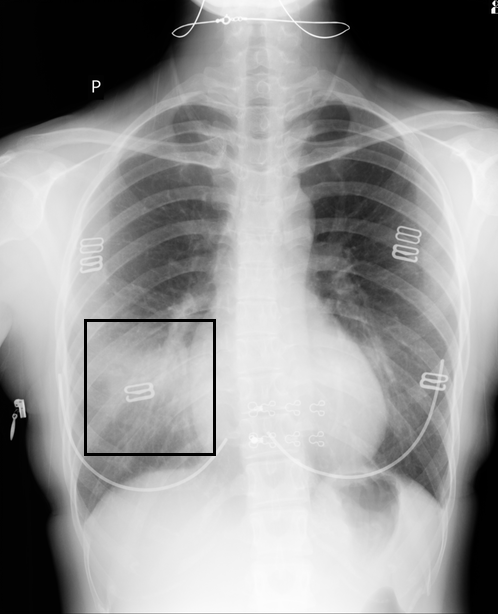} 
      \includegraphics[height=5cm]{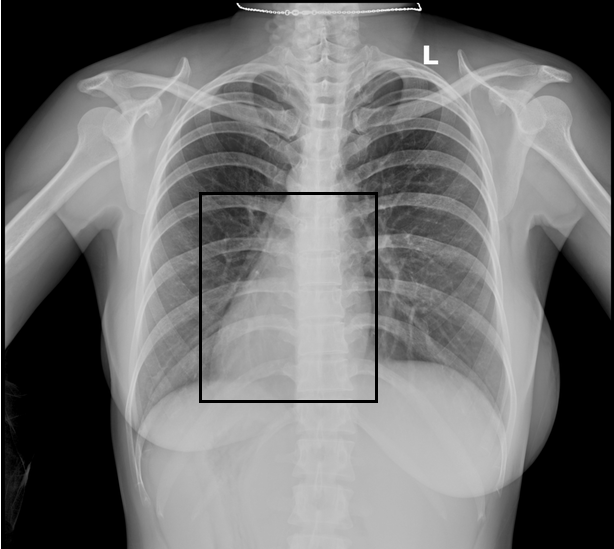} 
    \caption{Examples of lesions found on images checked by three radiologists and classified as No finding. The image on the left should be annotated as containing consolidation/pneumonia label, and the~image on the right as Other lesion (actually dextrocardia).}
    \label{fig:nofindings-error}
\end{figure}

\paragraph{\textbf{Lesions present on chests with ”no findings” label}} Our expert radiologist analyzed ten randomly selected images annotated by each of the~seventeen radiologists (R1-R17). It appeared that many annotations are missing. Surprisingly, we found out that although there was a~general consensus between dataset annotators when labeling "no findings", actually there are some anomalies that should be marked. The review result is presented in Table \ref{tab:no_f_rev}, and sample errors in Figure~\ref{fig:nofindings-error}.

\begin{table}[!h]

\centering
\caption{Our expert radiologist checked 10 images annotated by each radiologist R1-R17 as "no findings". The number of images wrongly annotated is presented below.\\}
    
    % {\scriptsize
    
    % \scalebox{1.0}{
    % \setlength{\tabcolsep}{2pt}
    
    % \hskip-0.2cm
    \begin{tabular}{|p{2.4cm}|p{0.25cm}|p{0.25cm}|p{0.25cm}|p{0.25cm}|p{0.25cm}|p{0.25cm}|p{0.25cm}|p{0.25cm}|p{0.25cm}|p{0.41cm}|p{0.39cm}|p{0.41cm}|p{0.39cm}|p{0.41cm}|p{0.39cm}|p{0.41cm}|p{0.39cm}|p{0.41cm}|} 
    \hline
    radiologist's ID & R1 & R2 & R3 & R4 & R5 & R6 & R7 & R8 & R9 & R10 & R11 & R12 & R13 & R14 & R15 & R16 & R17 \\ \hline
    number of errors & 0 & 4 & 0 & 2 & 1 & 1 & 1 & 3 & 1 & 1 & 2 & 5 & 5 & 1 & 1 & 1 & 3 \\ \hline
    \end{tabular}

    % }
    % }
    \label{tab:no_f_rev}

\end{table}

\begin{figure}[!h]
    \centering
      \includegraphics[height=6cm]{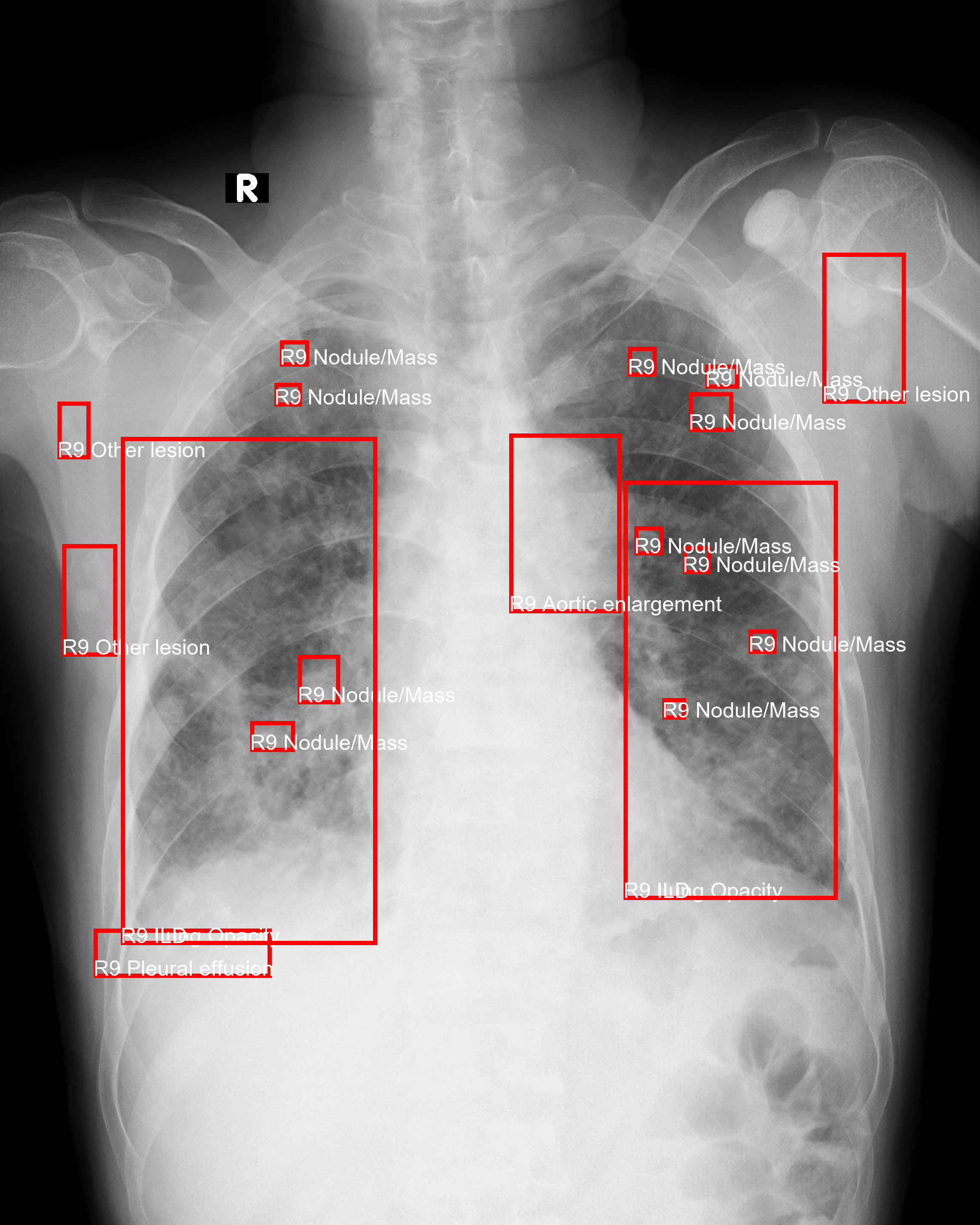} 
      \includegraphics[height=6cm]{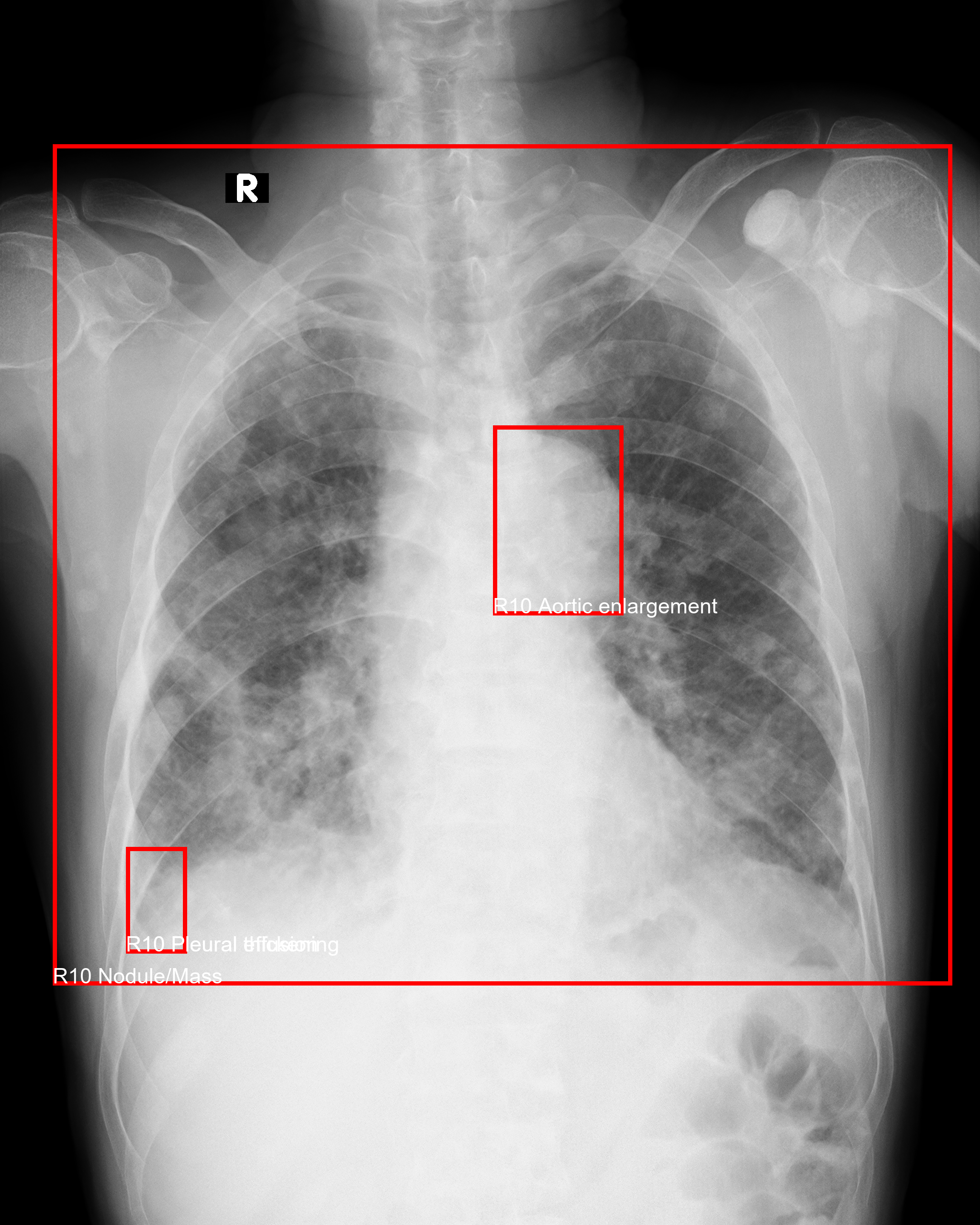} 
    \caption{Examples of inconsistency between radiologists related to the usage of a single box to mark many anomalies of the same class. On the~left image, there are two big boxes each for the left and the right lung and many small boxes, on the~right one, there is single box covering both lungs.}
    \label{fig:box-problem}
\end{figure}

\paragraph{\textbf{One bounding box for all lesions of the~same type, or one for each lesion.}} We found an~additional problem with not clear rules for annotations, some radiologists use a~single box to cover few anomalies, others mark each anomaly separately. Some example is presented in Figure \ref{fig:box-problem}. This results in model training problems as it introduces high noise in labels. 

Moreover, it influences a~model quality. The metric mAP at IoU 40, chosen for the~competition, means that predicted bounding box has to overlap with ground-truth box in at least 40\%. The problem is that if radiologists' annotations (ground truth) do not meet this requirement, how is it possible to train an~AI model with such noisy labels to get a~good result.

\paragraph{\textbf{Different procedure of preparing train and test sets}} The train and test sets were prepared differently. In both, the~annotations were made independently by three radiologists for each image. According to \cite{Nguyen2020}, in the~test set, there was an~additional processing step. The labels were additionally verified and a~consensus between two radiologists was reached.

The problem is that there are considerable differences between radiologists. One approach is to select only critical findings and discard other annotations as unnecessary, which is acceptable for radiologists, but very challenging for nowadays ML model architectures. Typically, there is an~assumption that a~ML model should be trained on data similar to the~target, and in order to deal with noise, more data is required.

The second issue is the~radiologist bias. From the~training set analysis, we found out that most annotations were made by actually three radiologists (R8-R10). However, it is not known whether images annotated by those were used in the~test dataset. This bias is reinforced by the~additional two radiologists who made a~consensus over annotations of three radiologists including standardization of label definitions. 

The role of two expert radiologists is unclear. It seems that those two only corrected annotations made by others. Their role should be much bigger, they are necessary to control if the~annotation rules are well understood, and to clarify them if a~new corner case arises. The standardized criteria for annotation should be prepared.

%------------------------------------------------------------------------

\subsection{Data quality}

\paragraph{\textbf{Missing or wrong metadata in DICOMs}}

In Figure \ref{fig:labels}, it is presented that some images suffer from missing data usually distributed in an~extensive DICOM header, either from the~lack of age or sex. The data is also classified as missing when the~type of data is incorrect (i.e., a~letter instead of a~number). 
68\% of the~observations do not have information about age, and 17\% about sex. The sex parameter is set to O (other) for 34\% of the~images. The rest of the~dataset is fairly balanced (M: 26\%, F: 23\%).
There is a~lot of instances where the~age is equal to 0 or is far greater than 100 (i.e. 238). This leaves us with only 25\% of images with valid ages between 1-99. 

\begin{figure}[!h]
    \centering
    \includegraphics[width=8cm]{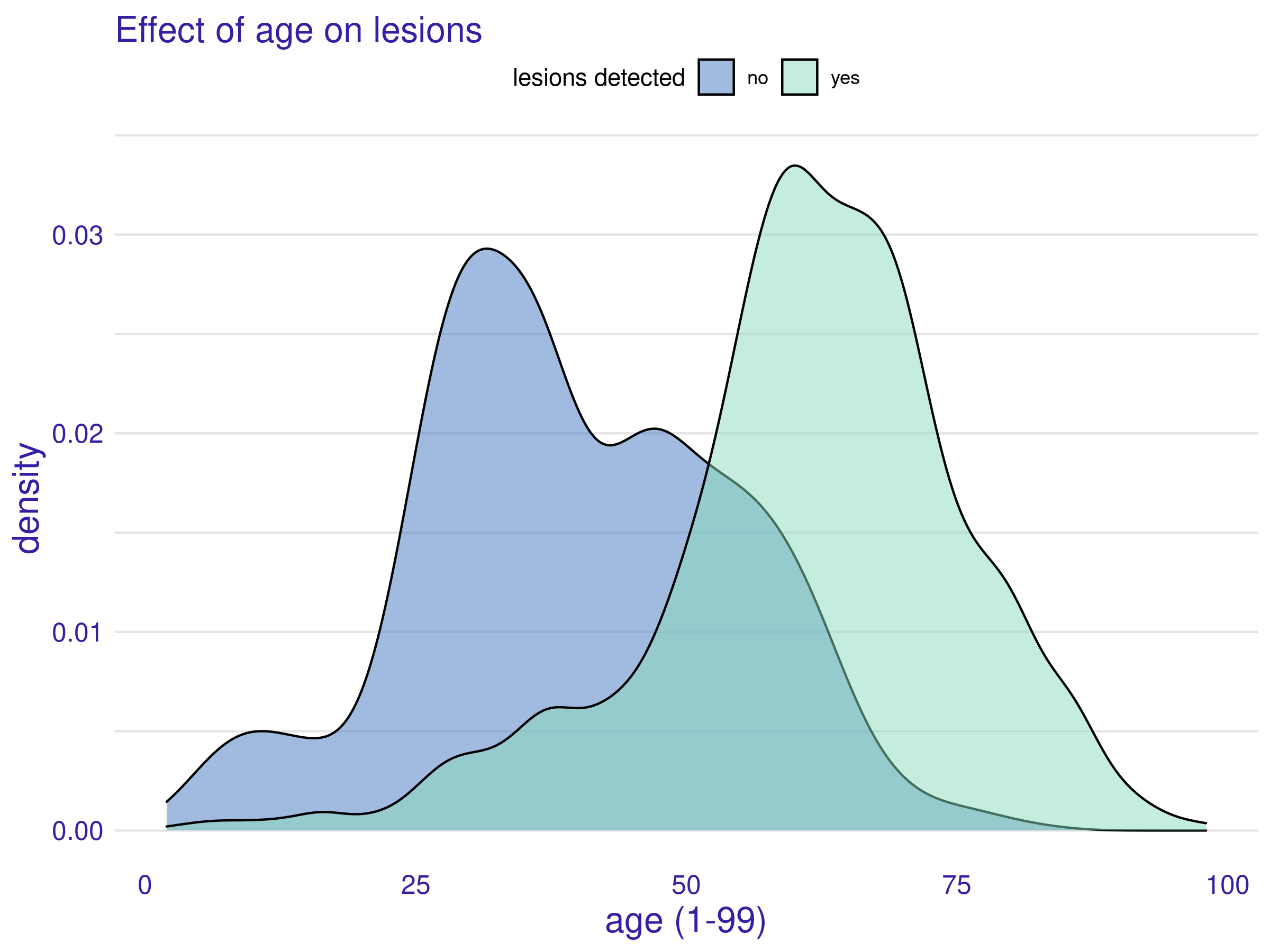}
    \caption{Density plots of age grouped by existence of an~illness. The probability of a~young person having a~lesion is lower than for the~older person.}
    \label{fig:age_illness}
\end{figure}

The lack of reliable information about age or sex is unfavorable because such attributes might be correlated with certain diseases, or having a~disease at all. For example, for younger people, the~probability of having lesions is significantly lower than for older people. It can be seen in Figure \ref{fig:age_illness} where density plots for ages of patients with and without detected diseases differ visibly.

\paragraph{\textbf{Children present in the~dataset}}
In the~training dataset, there are 107 images of children (ages 1-17). This might be a~problem as child anatomy is different from adults (i.e., shape of heart, mediastinum, and bone structure) and so are the~technical aspects of the~child's X-ray (position of hands) \cite{Hryniewska2020}. The model might recognize such relationships.

According to \cite{Nguyen2020}, pediatric X-rays should have been removed from the data during the data filtering step, but we found they were accidentally left.

As children are not small adults, they should be removed in order not to introduce additional noise during model training.

\paragraph{\textbf{Two monochromatic color spaces}}
Another valid concern is Photometric Interpretation, which specifies the~intended interpretation of the~image pixel data. Some images are of type \textit{monochrome1} (17\%) and some of \textit{monochrome2}. The difference is that in the~first case the~lowest value of a~pixel is interpreted as white and in the~second case as black. This may produce some inefficient models when not taken into consideration.

\begin{figure}[!h]
    \centering
      \includegraphics[height=5cm]{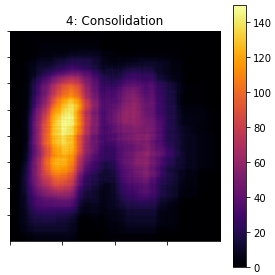} 
      \includegraphics[height=5cm]{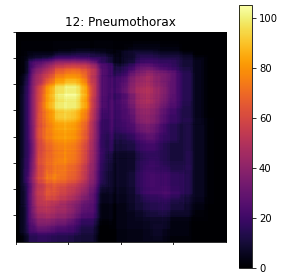}
    \caption{Examples of lesions (consolidation and pneumothrox) that should be present symmetrically in both parts of the~lungs.}
    \label{fig:alnom-heatmap}
\end{figure}

\paragraph{\textbf{Lesions localization imbalance}} There are fourteen annotated anomalies. In regular clinician practice, all except two (aortic enlargement, cardiomegaly) are distributed similarly on both lungs sides. There should be a~similar number of lesions in the~right lung as well as in the~left one. However, in Figure \ref{fig:alnom-heatmap}, we placed heatmaps that should show the~anomalies symmetrically appeared in both lungs. Before heatmaps were calculated, images from the~training set were centered.

\begin{figure}[!h]
    \centering
      \includegraphics[height=5cm]{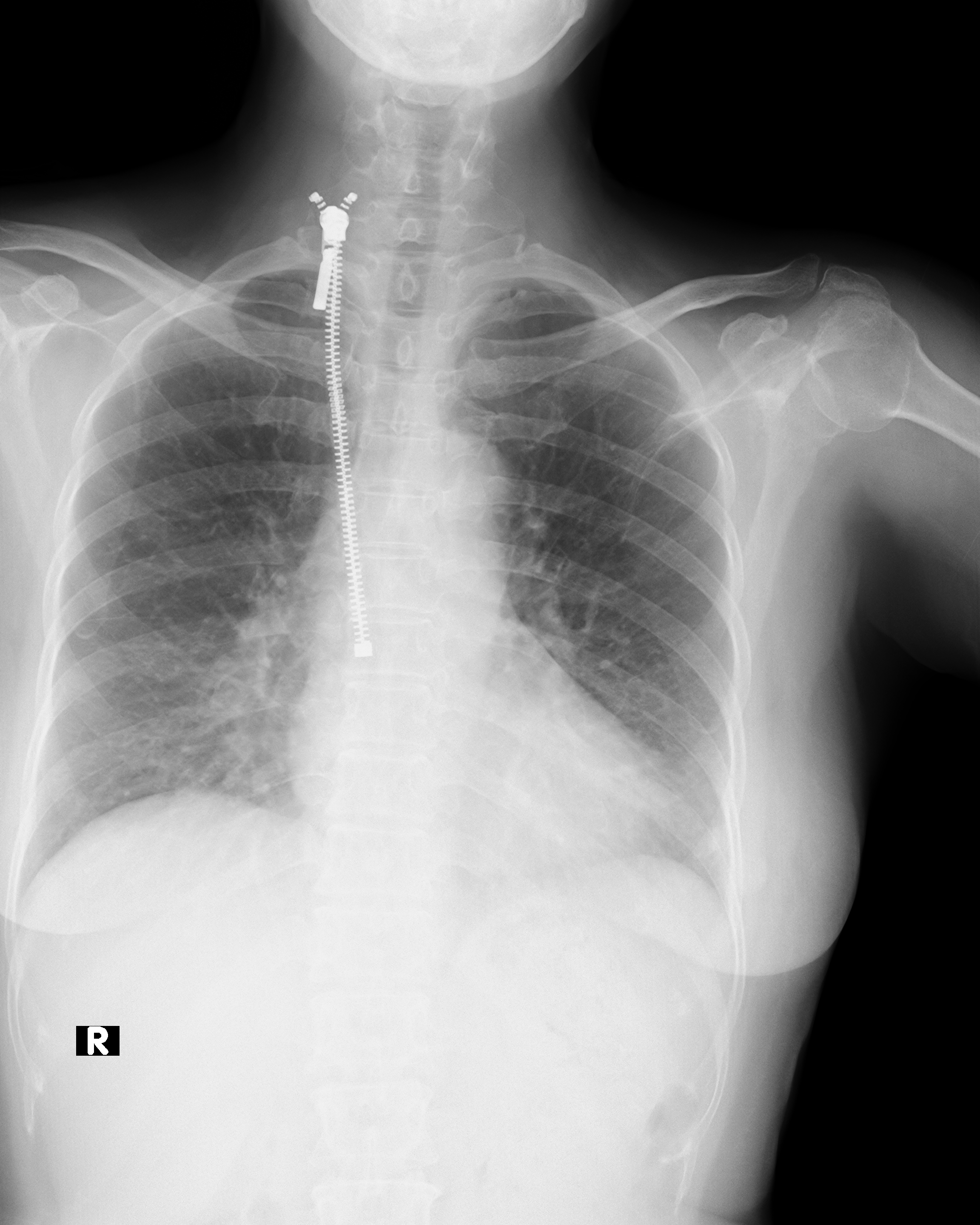} 
      \includegraphics[height=5cm]{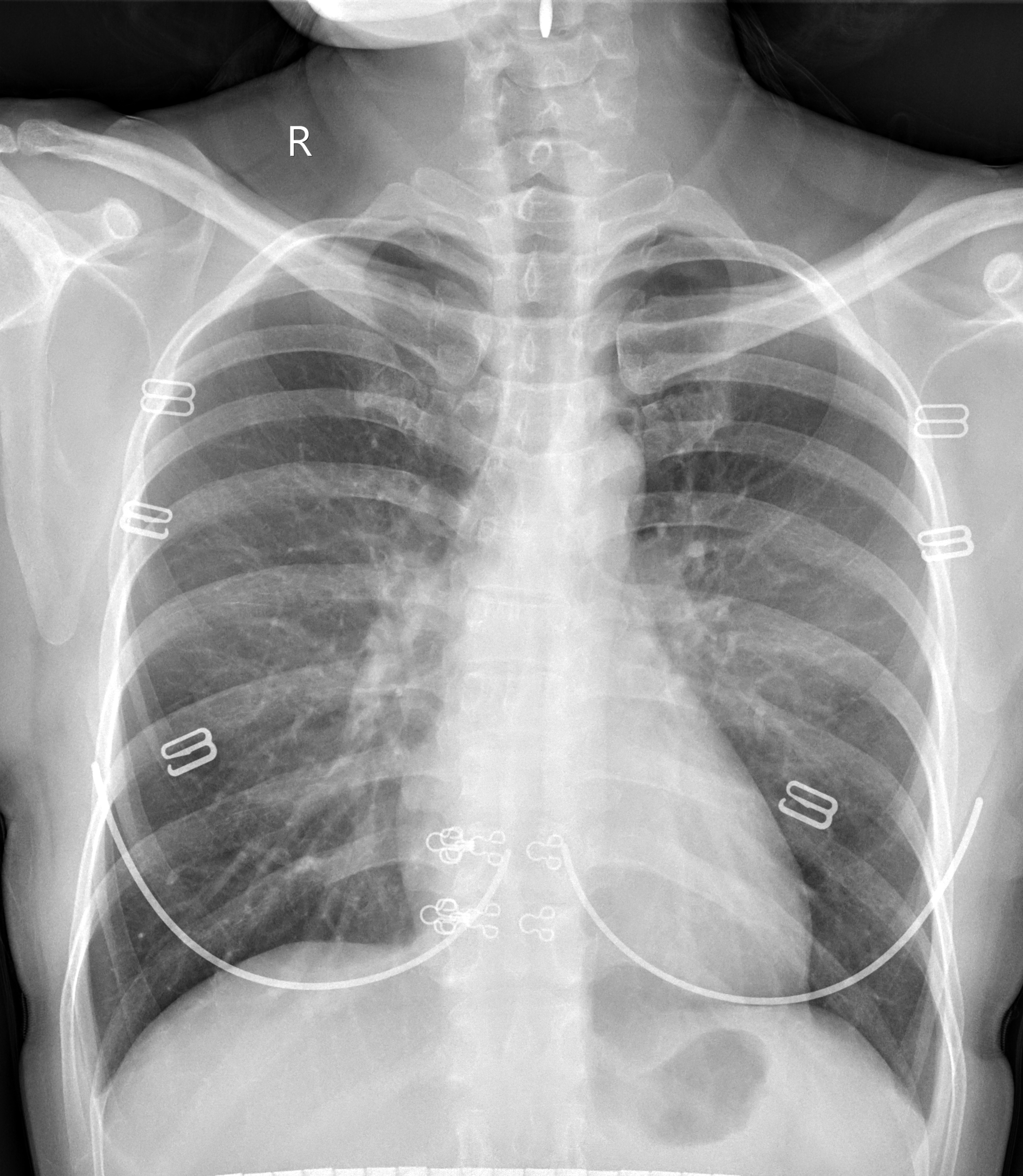} 
    \caption{Example of clothes artifacts.  From the~left, there are: a~zipper, a~bone in a~bra.}
    \label{fig:clothes}
\end{figure}

\paragraph{\textbf{Parts of clothes present in the~X-rays.}}
Undesirable artifacts are present in many images, which in some cases can reduce the~diagnostic value of the~image, and when used for machine learning, introduce additional noise. Some examples are shown in Figure \ref{fig:clothes}. These artifacts can be easily avoided during image acquisition, by asking the~patient to remove all parts of the~clothes that may influence X-ray imaging, for example, chains, bras, clothes with buttons, and zippers. If artifacts cannot be prevented, they can be removed during image preprocessing, before the image is shown to the model.

\paragraph{\textbf{Letters and annotations present in the~X-rays}} Letters and/or annotations present in some lung images should be removed during preprocessing to prevent a~neural network from learning those patterns. The model should learn how to differentiate labels by focusing on image features, not on descriptions in the~images.

%------------------------------------------------------------------------

\section{Model fairness concerns}
We created a Faster R-CNN model \cite{detectron2} with input image size 1024x1024 that detects regions with potential lesions with mAP at IoU $>$ 0.4 equal 18.1\%. To assess the~fairness of this model, we assumed that it is correct when it detects the~class anywhere in the~picture if the~class is present in any radiologist's annotations. Then, we performed 14 checks over all classes of lesions (simple binary split - whether the~illness was present among annotations or not). We checked popular fairness metrics over 2 features - age (missing, young($<$ 50 y.o.), and old ($\geq$ 50 y.o.)) and sex (Male, Female, Other, Missing). From this analysis, it became apparent that our model has problems with Predictive Parity (PPV, precision) over these subgroups. For example, for aortic enlargement among the~age, the~precision in subgroup young was 0.13 and in subgroup old 0.74. It is essential to evaluate the~model in this way to be aware of its faults or to try to mitigate the~potential bias.  

%------------------------------------------------------------------------

\section{Conclusion}
The quality of a~model is inherently bound to the~quality of the~data on which it is trained. Development of a~reliable model should begin with data acquisition and annotation. At the~model development stage, we cannot make the~model fulfill all responsible AI and fairness rules if the~data and their annotations are of insufficient quality.

%Bibliography
\bibliographystyle{unsrt}  
\bibliography{references}

\end{document}